\documentclass{article} 
\usepackage{nips14submit_e,times}
\usepackage{hyperref}
\usepackage{url}
\usepackage{graphicx}
\usepackage{amssymb}
\usepackage{natbib}
\usepackage{subcaption}
\usepackage{algorithm,algorithmic}
\setcitestyle{authoryear,open={(},close={)}}

\graphicspath{ {images/} }

\title{KF-LAX: Kronecker-factored curvature estimation for control variate optimization in reinforcement learning}

\author{
Mohammad Firouzi \\
University of Toronto \\
\texttt{firouzi@cs.toronto.edu} \\
}

\nipsfinalcopy 

\begin{document}

\maketitle

\begin{abstract}
A key challenge for gradient based optimization methods in model-free reinforcement learning is to develop an approach that is sample efficient and has low variance. In this work, we apply Kronecker-factored curvature estimation technique (KFAC) to a recently proposed gradient estimator for control variate optimization, RELAX, to increase the sample efficiency of using this gradient estimation method in reinforcement learning. The performance of the proposed method is demonstrated on a synthetic problem and a set of three discrete control task Atari games. \let\thefootnote\relax\footnote{In Neurips 2018 Workshop on Deep Reinforcement Learning}
\end{abstract}

\section{Introduction}

Latent variable models are widely used in machine learning. Specifically, discrete latent variable models have been studied in many areas such as reinforcement learning (RL) and natural language processing. In the RL applications, for the optimization of both continuous and discrete variable models, policy gradient methods \citep{Williams:1992:SSG:139611.139614, glynn1990likelihood} have gained high attention and have been applied in a wide variety of tasks. However, they mostly suffer from having a high variance for the gradient estimation. 

Various works have focused on reducing the variance of the estimated gradient in the policy gradient methods. A subset of these approaches try to reduce the variance through use of a well designed control variate or a baseline. For example, Q-prop \citep{gu2016q} and MuProp \citep{gu2015muprop} use control variate based on the first order Taylor expansion of the target function around a fixed point, and \citep{wu2018variance} uses a factorized action-dependent baseline. In many applications, despite the estimated gradient being unbiased, it still suffers from high variance even when it is aligned with a control variate. In cases that the target function is continuous, reparameterization trick \citep{kingma2013auto, rezende2014stochastic} could be used as a way of obtaining a low variance gradient estimator. 

Particularly, in discrete domains, concrete relaxation \citep{maddison2016concrete, jang2016categorical} uses this technique to obtain a low variance but biased gradient estimator. More recent gradient estimators such as REBAR \citep{tucker2017rebar} and RELAX \citep{grathwohl2017backpropagation} leverage both the control variate and the reparameterization trick to obtain a low variance and unbiased estimator. To estimate the gradient and apply the reparameterization trick, REBAR uses two different uniform random variables. In general, REBAR is an unbiased gradient estimator. RELAX gradient estimation approach is further discussed in the next sections.

Natural gradient descent is an approach that leverages the local curvature structures of a target function and can provide gradient updates invariant to the parameterization. In problems that parameter space has an underlying structures, the ordinary gradient does not represent the steepest descent direction, but the natural gradient does \citep{amari1998natural}. Therefore, an efficient estimation of the natural gradient can provide a practical way to build a faster training procedure. In particular, in reinforcement learning, natural policy gradient method \citep{kakade2002natural} was introduced over a decade ago, and works such as \citep{wu2017scalable} and \citep{schulman2015trust} successfully applied it in practice for discrete and continuous control tasks. As an another practical usage of the natural gradient, authors in \citep{hoffman2013stochastic} used a noisy computation of natural gradient to update variational parameters in mean field variational inference, and applied it for topic modeling of large data sets.

KFAC \citep{martens2015optimizing} is a technique to approximate the local curvature structure of a given function with an efficient way of using mini-batch samples of data. Sample efficiency and fast optimization procedure are critical challenges in deep RL. As a second order optimization technique, leveraging KFAC to estimate the natural gradient can be helpful to train an agent faster. In this paper, inspired by KFAC and a recently introduced gradient estimator, RELAX \citep{grathwohl2017backpropagation}, we develop a gradient optimization procedure for black box functions which approximates the natural gradient for a differentiable surrogate function as a control variate. Black box functions are particularly important in RL where the reward function is not necessarily known. 
Section 2 describes the preliminaries. Section 3 is devoted for the proposed method. Section 4 provides a summary of related works. The proposed method is applied to a synthetic problem and a few discrete control task Atari games in section 5, and the conclusion and future work are provided in the last sections.

\section{Preliminaries}
Many loss functions in machine learning such as variational objective in variational inference, expected reward in reinforcement learning, and empirical risk in supervised learning can be written as an expectation of a function over the data distribution. Consider the optimization problem of a function in the form of $\mathbb{E}_{p(x|\theta)}[f(x)]$ with respect to $\theta$. Some of the popular techniques to obtain a gradient estimator for the parameters of this loss function are described in sections 2.1-2.3. In section 2.4 the natural gradient and KFAC are explained in more details.

\subsection{Score function estimator}
A general approach to estimate the gradient of $\mathbb{E}_{p(x|\theta)}[f(x)]$ with respect to $\theta$ is known as the score function estimator (also known as REINFORCE, and likelihood ratio estimator) \citep{Williams:1992:SSG:139611.139614,glynn1990likelihood} which uses the derivatives of $\log p(x|\theta)$ to estimate the gradient as,
$$
f(x) \nabla_\theta \log p(x|\theta)
$$
Where $x$ is sampled from $p(x|\theta)$. A straight forward calculation shows that the score function is an unbiased gradient estimator \citep{Williams:1992:SSG:139611.139614}. Moreover, this estimator assumes no special restriction for the function $f$ and is generally applicable for black box functions. In situations that $f$ directly depends on $\theta$ and the dependency is known, the derivatives of $f$ can be added to the above estimator to come up with an unbiased estimator. 

\subsection{Reparameterization trick}
Reparameterization trick \citep{kingma2013auto, rezende2014stochastic} is a technique for cases that $f$ is a continuous and differentiable function, and $x$ can be written as a continuous function of $\theta$ and a random noise $\epsilon$ with a known distribution. The reparameterized gradient estimator has the form,
$$
\quad \quad \quad \quad \quad \quad \quad \quad \quad \quad \frac{\partial f}{\partial T} \frac{\partial T}{\partial \theta}, \quad x = T(\epsilon, \theta) ,\epsilon \sim p(\epsilon)
$$
Reparameterization gradient estimate is generally unbiased and has low variance. In practice, the standard Gaussian and Gumbel distributions are popular distributions for reparameterizing the continuous and discrete $x$ values, respectively \citep{maddison2016concrete, rezende2014stochastic, kingma2013auto, jang2016categorical}.

\subsection{Control variates}
For a given gradient estimator $\hat{g}$, using a well designed function $c(x)$ which $\mathbb{E}_{p(x|\theta)}[c(x)]$ can be calculated analytically can result a lower variance gradient estimator without changing the bias of the initial estimator as follow,
$$
\hat{g}_{new}(x) = \hat{g}(x) - c(x) +  \mathbb{E}_{p(x|\theta)}[c(x)]
$$
Stronger positive correlation between $c(x)$ and $\hat{g}(x)$ results more reduction in the variance of the new estimator.

\begin{algorithm}[t]
\caption{KF-LAX}
\begin{algorithmic}
\label{alg1}
\REQUIRE $f(.), \log p(x|\theta), x=T(\theta,\epsilon), p(\epsilon), $ a neural network surrogate $c_\phi(.)$ with weights $W_l$ for layer $l, $ step sizes $\alpha_1, \alpha_2$,
\WHILE{not converged}
\STATE $\epsilon \sim p(\epsilon)$
\STATE $x \leftarrow T(\theta, \epsilon)$
\STATE $\hat{g}_\theta \leftarrow [f(x) - c_\phi(x)] \frac{\partial}{\partial \theta} \log p(x|\theta) + \frac{\partial}{\partial \theta} c_\phi(x)$
{
\FOR{each layer $l$ of $c_\phi$}
\STATE Estimate the matrices $A$ and $S$ for the layer $l$ using KFAC.
\STATE $W_l \leftarrow W_l - \alpha_2 A^{-1} \nabla_{W_l} \hat{g}_\theta^2 S^{-1}$
\ENDFOR
}
\STATE $\theta \leftarrow \theta - \alpha_1 \hat{g}_\theta$
\ENDWHILE
\RETURN $\theta$
\end{algorithmic}
\end{algorithm}

\subsection{Natural gradient and Kronecker-factored curvature approximation}
Consider the problem of minimizing $J(\theta)$ with respect to $\theta \in \Theta$. A Conventional gradient descent approach uses Euclidean distance over the parameter space $\Theta$. However, Euclidean distance does not benefit from the curvature information of the function and may result a slower convergence. Natural gradient \citep{amari1998natural} uses particular Mahalanobis distance as a local metric and gives an update rule of the form $\Delta \theta \propto -F^{-1} \nabla_\theta J$, where $F$ is the Fisher information matrix which is a function of $\theta$.

Modern neural networks could have millions of parameters. In cases that the optimization is with respect to a neural network parameters, computing the Fisher information matrix and its inverse can be computationally impractical and inefficient. For example, for a neural network with only one fully-connected layer with 1000 input and 1000 outputs, direct estimation of the entries of the Fisher information matrix which is a square matrix with a million rows might be inefficient. Kronecker-factored approximate curvature (KFAC) \citep{martens2015optimizing} leverages the properties of Kronecker product to approximate the Fisher information matrix and its inverse efficiently. Consider a multi-layer neural network. Let the output of the neural network be $p(y|x)$ where $x$ denotes the input, and let $L=\log p(y|x)$ indicates the log likelihood. For the $l$-th layer of the neural network, let $d_{in}$, $d_{out}$ denote the input and output size of the layer (the index $l$ is dropped for simplicity), $W_l \in \mathbb{R}^{d_{in}\times d_{out}}$ the weights, $a \in \mathbb{R}^{d_in}$ input vector to the layer, and $s = W_l a$ as the pre-activation vector for the next layer. We note that $\nabla_{W_l} L = (\nabla_{s} L) a^T$. KFAC uses this relationship to approximate the Fisher information matrix in a block diagonal manner, where each block corresponds to a layer. For the layer $l$, denote the corresponding block as $F_l$. The approximate Fisher information for this layer can be written as,
$$
F_l = \mathbb{E} [vec\{ \nabla_{W_l} L\}vec\{ \nabla_{W_l} L\}^T] = \mathbb{E} [aa^T \otimes \nabla_{s} L (\nabla_{s} L)^T]$$
$$\approx \mathbb{E} [aa^T] \otimes \mathbb{E} [\nabla_{s} L (\nabla_{s} L)^T] := A \otimes S := \hat{F}_l
$$
Where $A := \mathbb{E} [aa^T]$, and $S := \mathbb{E} [\nabla_{s} L (\nabla_{s} L)^T]$. $A$ and $S$ can be approximated during training using mini-batches of the data, or through the states and actions resulting from the agent experience of different trajectories. Having these approximations, the inverse of the block could be obtained using the following property of Kronecker product, 
$$(X \otimes Y)^{-1} = X^{-1} \otimes Y^{-1}$$
Therefore, the natural gradient update for the layer $l$ can be approximated as,
$$\Delta W_l = \hat{F_l}^{-1} \nabla_{W_l} J = A^{-1} \nabla_{W_l} J S^{-1}$$
For more details about the derivation refer to the \citep{martens2015optimizing}.

\begin{figure}[t]
\centering
\begin{subfigure}[t]{0.48\textwidth}
        \centering
        \includegraphics[width=5.5cm]{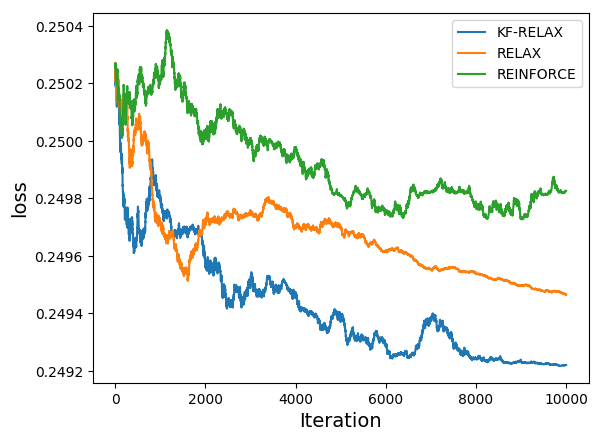}
\end{subfigure}
\begin{subfigure}[t]{0.48\textwidth}
        \centering
        \includegraphics[width=5.3cm]{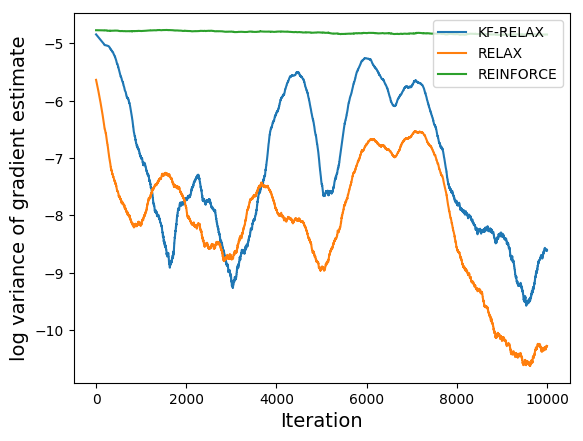}
\end{subfigure}
~
\caption{Loss and variance curve of the REINFORCE, RELAX and KF-RELAX gradient estimators in the synthetic problem with $t=0.499$. The learning rate for optimizing $\theta$ is set to a common value for all the estimators. The surrogate function learning rate is set to a same value for both the RELAX and KF-RELAX.}
\label{fig:TOY}
\end{figure}

\section{Proposed Method}
This section uses KFAC for the LAX/RELAX \citep{grathwohl2017backpropagation} gradient estimators which can be applied in both continuous and discrete domains. LAX (continuous cases) and RELAX (discrete cases) leverage the score function and the reparameterization trick to obtain a low variance estimator without adding bias to the initial gradient estimator.

Let $x$ be a continuous random variable and consider the optimization of $\mathbb{E}_{p(x|\theta)}[f(x)]$ with respect to $\theta$. The LAX estimator \citep{grathwohl2017backpropagation} is given by,
$$
\hat{g}_{LAX} = \hat{g}_{REINFORCE}[f] - \hat{g}_{REINFORCE}[c_\phi] + \hat{g}_{REPARAM}[c_\phi]
$$
$$
= [f(x) - c_\phi(x)] \frac{\partial}{\partial \theta} \log p(x|\theta) + \frac{\partial}{\partial \theta} c_\phi(x), \quad \quad x=T(\theta, \epsilon), \epsilon \sim p(\epsilon)
$$
where $c_\phi$ is the surrogate function. Parameters of the surrogate function are trained in order to minimize the variance of the gradient estimator. In case that that $c_\phi$ is equal to $f$, LAX and the reparameterization gradient estimator are the same. By applying KFAC to estimate the natural gradient for the surrogate function parameters, KF-LAX procedure is obtained. The procedure is described in the algorithm \ref{alg1}.

Similarly, in case that $x$ is a discrete random variable, RELAX can be written as,
$$
\hat{g}_{RELAX} = [f(x) - c_\phi(\tilde{z})]\frac{\partial}{\partial \theta} \log p(x|\theta) + \frac{\partial}{\partial \theta} c_\phi(z) - \frac{\partial}{\partial \theta} c_\phi(\tilde{z}),
$$
$$
x = H(z), z \sim p(z|\theta), \tilde{z} \sim p(z|x, \theta)
$$
where $H$ represents the heaviside function\footnote{$H(z)=1$ if $z \geq 0$ and $H(z)=0$ if $z < 0$}. Following the same approach for the LAX, we get another optimization algorithm for RELAX which we call KF-RELAX. This algorithm is provided in Appendix \ref{kf-relax}.

\section{Experiments}
\subsection{Synthetic problem}
\label{toy_prob}
KF-RELAX is applied to a synthetic problem. Consider the minimization of $\mathbb{E}_{p(b|\sigma(\theta))}[(b-t)^2]$ where $t \in (0,1)$ is a constant and $b$ is a binary random variable sampled from a Bernoulli distribution with parameter $\sigma(\theta)$, where $\sigma(.)$ indicates the sigmoid function. According to the selection of $t$, the optimal distribution is a deterministic Bernoulli distribution with parameter $0$ or $1$. Figure \ref{fig:TOY} compares the convergence of REINFORCE, KF-RELAX and RELAX for $t=0.499$, using only a single sample of $x$ per iteration. Also, in appendix \ref{unbiased_rlx_kfrlx}, it is empirically shown that all gradient estimators are unbiased.

\subsection{Reinforcement learning}
Consider a finite discrete-time Markov Decision Process (MDP) which is a tuple $(\chi, \rho_0, \gamma ,A, r, P)$. At time $t=0$, an agent is in the state $s_0$ which is sampled from the initial distribution $\rho_0: \chi \rightarrow \mathbb{R}$. At each time $t$, the agent chooses an action $a_t \in A$ from its policy distribution $\pi_\theta(a_t|s_t)$, gets a reward according to the reward function $r(s_t, a_t)$ and transitions to the next state $s_{s+1}$ according to the Markov transition probabilities $P(s_{t+1}|s_t, a_t)$. Final goal of the agent is to maximize its expected reward $J(\theta)= \mathbb{E}_{s_0, a_0, \tau}[\sum_{t=0}^{\infty}\gamma^{t} r(s_t, a_t)] = \mathbb{E}_{s\sim \rho^\pi(s), a, \tau}[\sum_{t=0}^{\infty}\gamma^{t} r(s_t, a_t)]$, where $\tau$ indicate the trajectories, $\gamma$ is the discount factor and $\rho^\pi = \sum_{t=0}^{\infty} \gamma^{t} P(s_t = s)$ is the unnormalaized state visitation frequency. Using the policy gradient theorem \citep{sutton1998reinforcement}, the gradient of $J(\theta)$ can be written in the following expectation form,
$$
\nabla J(\theta) = \mathbb{E}_{s\sim \rho^\pi(s), a, \tau}[Q^\pi(s,a) \nabla_\theta \log \pi_\theta(a|s)] = \mathbb{E}_{s\sim \rho^\pi(s), a, \tau}[A^\pi(s,a) \nabla_\theta \log \pi_\theta(a|s)]
$$

Where, $Q^\pi(s,a)$ is the state-action value function, $V^\pi(s)$ is the value function, and $A^\pi(s,a) = Q^\pi(s,a) - V^\pi(s)$ is the advantage function. Due to simplicity of the selected tasks, same as \citep{grathwohl2017backpropagation}, we use the $Q$ function approximation, instead of directly estimating the advantage function.

\subsubsection{Discrete Control Tasks}
KF-RELAX and RELAX performance are compared in three discrete control RL tasks. According to appendix C of the \citep{grathwohl2017backpropagation}, using a sampled trajectory, the RELAX gradient estimator can be written as,

$$
\hat{g}^{RL}_{RELAX} = \sum_{t=1}^{T} \frac{\partial}{\partial \theta} \log \pi(a_t|s_t, \theta) [\hat{Q}^\pi(a_t,s_t) -c_\phi(\tilde{z}_t,s_t)] - \frac{\partial}{\partial \theta} c_\phi(\tilde{z}_t,s_t) + \frac{\partial}{\partial \theta} c_\phi(z_t,s_t)
$$

KF-RELAX gradient estimator has a similar form with a difference that at each step, the natural gradient of the surrogate function is estimated.

\begin{figure}
\centering
\begin{subfigure}[t]{0.3\textwidth}
        \centering
        \includegraphics[width=4.5cm]{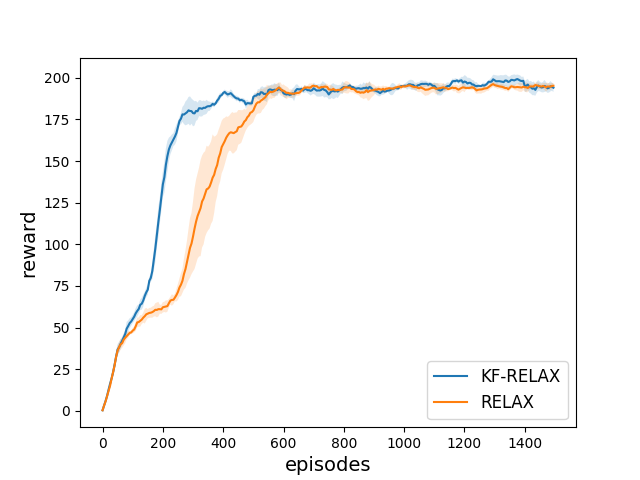}
        \caption{Cart pole}
\end{subfigure}
~
\begin{subfigure}[t]{0.3\textwidth}
        \centering
        \includegraphics[width=4.5cm]{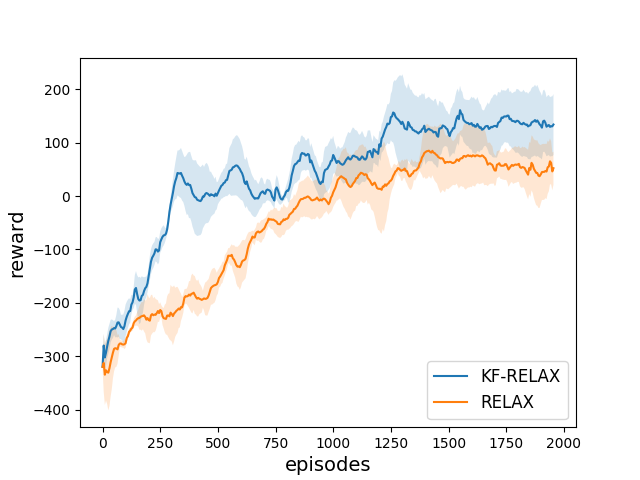}
        \caption{Lunar lander}
\end{subfigure}
~
\begin{subfigure}[t]{0.3\textwidth}
        \centering
        \includegraphics[width=4.5cm]{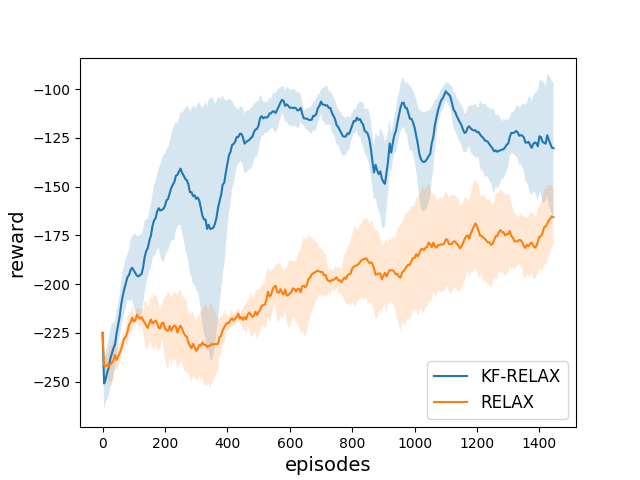}
        \caption{Acrobot}
\end{subfigure}
\caption{Training curve for KF-RELAX and RELAX on three discrete task Atari games.}
\label{fig:rl_res}
\end{figure}

The learning rate for both the agent and the surrogate function were chosen from the set $\{0.03, 0.01, 10^{-3}, 10^{-4}\}$. $c_\phi$ is chosen as a three fully-connected layer neural network. For the RELAX, all models were trained using Adam \citep{kingma2014adam}. Entropy regularization with a weight of $0.01$ was used to increase the exploration, and the discount factor was set to $0.99$. Tikhonov damping technique described by \citep{martens2015optimizing} with the value in $\{0.1, 0.01, 10^{-3}, 5\times 10^{-4}\}$ was used to further control the training stability. The same value of mini-batch size was used for both gradient estimators. In addition, as in \citep{wu2017scalable}, a trust region approach was used for KF-RELAX, whereby the parameter updates capped at a specific upper bound to prevent the agent from converging to a near-deterministic policy during the early training stages. The value of the upper bound was chosen from $\{10^{-3}, 10^{-4}, 10^{-5}, 10^{-6} \}$. The experiments are done on Cart-pole, Lunar-Lander, and Acrobot which were selected from the OpenAI gym environments \citep{brockman2016openai}. The comparison of KF-RELAX and RELAX in these environments are shown in the figure \ref{fig:rl_res}.

\section{Related Works}
Natural policy gradient method was first introduced in \citep{kakade2002natural}. In particular case where the state-value function is approximated with a compatible function approximator, the natural gradient tends toward the optimal greedy policy improvement (Appendix \ref{compatible_adv_approx} shows the same fact holds for the advantage function which is frequently used for policy optimization). Natural gradient was used for actor-critic methods by \citep{peters2008reinforcement}. More recently, Trust Region Policy Optimization (TRPO) \citep{schulman2015trust} performs an optimization for each step of the policy update and the natural gradient policy update is a particular case of TRPO. However, TRPO does not scale to neural networks with larger architectures.

Kronecker-factored curvature approximation (KFAC) technique is also applied in practice to improve the training convergence speed. For example, \citep{grosse2016kronecker} used KFAC to estimate the Fisher information of the convolutional layers in neural networks. Recently, \citep{martens2018kronecker} extended KFAC to recurrent neural networks. \citep{wu2017scalable} applies KFAC to optimize both the actor and the critic of an actor-critic model for deep RL and proposed the ACKTR method. Inspired by ACKTR, this work uses the KFAC technique for the surrogate function in RELAX. In the continuous deep RL applications, a vanilla Gauss-Newton method can be used to leverage KFAC for the agent as well \citep{wu2017scalable, nocedal2006nonlinear}.

\section{Conclusion and Future Work}
Even though the KF-LAX (KF-RELAX) could provide stronger update rules for optimization and reduce the number of samples for convergence, tuning the hyperparameters for KFAC can be challenging in the deep RL framework as the estimation of the gradient may exhibit a high variance. Ill specification of parameters like the damping factor, and the scaling factor may cause a training collapse or no improvement in sample size needed for training. Fortunately, these hyperparameters are studied for practical usage and these studied are beneficial for the tuning process \citep{martens2015optimizing, wu2017scalable}.


Choosing a complicated architecture for the surrogate function can negatively influence the training progress or even add to the variance of the gradient estimation, hence degrading the convergence. In our experiments, a surrogate function with more than three layers increases the gradient estimation variance. This is considered a limitation of using this approach. Stabilizing the training process for more complicated surrogate functions is considered as a future work.

In this work we leveraged two recently introduced optimization techniques, KFAC and RELAX, to get more powerful gradient steps. We experimented the proposed approach on a few discrete control tasks in reinforcement learning. We believe this approach can be promising in improving the sample efficiency of RELAX in reinforcement learning. Further experiments for discrete and continuous tasks are considered as the future works.



\bibliography{ref}

\appendix
\section{KF-RELAX}
\label{kf-relax}

Using the KFAC technique for the RELAX gradient estimator results the KF-RELAX procedure for discrete cases which is explained in the Algorithm \ref{kf_rlx}. 

\begin{algorithm}[H]
\caption{KF-RELAX}
\label{kf_rlx}
\begin{algorithmic}[1]
\REQUIRE $f(.), \log p(x|\theta), x=T(\theta,\epsilon), p(\epsilon), $ a neural network surrogate $c_\phi(.)$ with weights $W_l$ for each layer $l, z=S(\theta, \epsilon), \tilde{z}=S(\theta, \epsilon|x)$  step sizes $\alpha_1, \alpha_2$, 
\WHILE{not converged}
\STATE $\epsilon, \tilde{\epsilon} \sim p(\epsilon)$
\STATE $z \leftarrow S(\theta, \epsilon)$
\STATE $x \leftarrow H(z)$
\STATE $\tilde{z} \leftarrow S(\theta, \epsilon|x)$
\STATE $\hat{g}_\theta \leftarrow [f(x) - c_\phi(\tilde{z})] \nabla_\theta \log p(x|\theta) + \nabla_\theta c_\phi(z) - \nabla_\theta c_\phi(\tilde{z})$
{
\FOR{each layer $l$ of $c_\phi$}
\STATE estimate the matrices $A$ and $S$ for the layer $l$ using  KFAC.
\STATE $W_l \leftarrow W_l - \alpha_2 A^{-1} \nabla_{W_l} \hat{g}_\theta^2 S^{-1}$
\ENDFOR
}
\STATE $\theta \leftarrow \theta - \alpha_1 \hat{g}_\theta$
\ENDWHILE
\RETURN $\theta$
\end{algorithmic}
\end{algorithm}


$$
$$
\section{RELAX and KF-RELAX are unbiased}
\label{unbiased_rlx_kfrlx}
\begin{figure}[H]
\centering
\includegraphics[width=5.5cm]{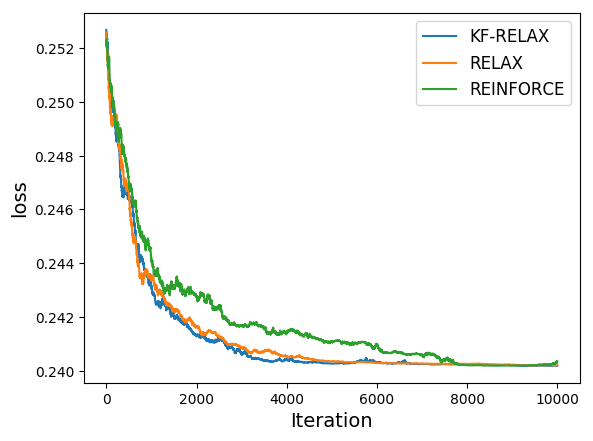}
\caption{Convergence of REINFORCE, RELAX, and KF-RELAX to the true solution in the synthetic problem with the target value $t=0.49$.}
\label{all_unbiased}
\end{figure}

Figure \ref{all_unbiased} shows that same as the REIFNORCE, RELAX and KF-RELAX are unbiased and converge to the true optimal solution in the synthetic problem which is described in section \ref{toy_prob}. Since for $t=0.499$ the REINFORCE gradient estimator does not necessarily converge, a simpler case $t=0.49$ is used in figure \ref{all_unbiased}.

\section{Use of the advantage function in the natural policy gradient}
\label{compatible_adv_approx}



In this section we rewrite a similar results to \citep{kakade2002natural} that with an approximation of the advantage function by some compatible function approximator, the natural gradient update rule tends to move toward the \textit{best} action.

\textbf{Lemma}: Suppose an approximation of the advantage function with a function $f^{\pi}(s,a;w)$ is desired. Let $f^\pi$ be in the form of $w^T \nabla \log \pi(a|s, \theta)$ and let $\tilde{w}$ be the minimizer of the square error $\epsilon(w,\pi) = \mathbb{E}_{s\sim \rho^{\pi},a} [(A^\pi(s,a) - f^{\pi}(s,a;w))^2]
$. Then
$$
\tilde{w} = \tilde{\nabla} J(\theta)
$$
where $J(\theta) = \mathbb{E}_{s,a,\tau}[\sum_{t=0}^\infty \gamma^t r_t]$, and $\tilde{\nabla}$ indicates the natural gradient.

\textbf{Proof}: Since $\tilde{w}$ minimizes the $\epsilon(w,\pi)$, we have $\frac{\partial \epsilon}{\partial w_i}=0$. For simplicity in the notation we set $\psi(s,a)=\nabla \log \pi(a|s, \theta)$. Therefore,

$$
\sum_{s,a} \rho^\pi(s) \pi(a|s,\theta) \psi(s,a) (\psi(s,a)^T\tilde{w} - A^\pi(s,a)) = 0
$$

Rearranging this equation,
$$
\sum_{s,a} \rho^\pi(s) \pi(a|s,\theta) \psi(s,a) \psi(s,a)^T \tilde{w} = \sum_{s,a} \rho^\pi(s) \pi(a|s,\theta) \psi(s,a) A^\pi(s,a)
$$

The policy gradient theorem \citep{sutton1998reinforcement} states that
$$\nabla J(\theta) = \mathbb{E}_{s\sim \rho^\pi(s), a, \tau}[Q^\pi(s,a) \nabla_\theta \log \pi_\theta(a|s)] = \mathbb{E}_{s\sim \rho^\pi(s), a, \tau}[A(s,a) \nabla \log \pi(a|s)]$$

Also, by using the fact that $\nabla \pi(a|s,\theta) = \pi(a|s,\theta) \psi^\pi(s,a)$ and the definition of the Fisher information matrix $F(\theta)= \sum_{s,a} \rho^\pi(s) \pi(a|s,\theta) \psi(s,a) \psi(s,a)^T$ both sides can be rewritten as,

$$
F(\theta)\tilde{w} = \nabla J(\theta)
$$

Solving for $\tilde{w}$ gives $\tilde{w} = F^{-1}(\theta) \nabla J(\theta)$ which is the definition of the natural gradient. \hfill $\square$


The following lemma states that based on our approximation of the $A^\pi(a|s,\theta)$, the natural gradient locally tends toward the \textit{best} action. An action $a$ is best if $a \in  argmax_{a'} f^\pi(s,a';\tilde{w})$. Intuitively, the best action is defined by performing greedy policy with regard to our approximation of the advantage function.

\textbf{Lemma}: Let $\theta' = \theta + \Delta \theta$ be the natural gradient update of the policy parameters where $\Delta \theta = \alpha \tilde{\nabla} J(\theta) = \alpha \tilde{w}$, then
$$
\pi(a|s,\theta') = \pi(a|s,\theta)(1+ f^\pi(s,a;w)) + O(\alpha^2)
$$
\textbf{Proof:} To the first order Taylor approximation, 
$$
\pi(a|s,\theta') = \pi(a|s,\theta) + \frac{\partial \pi(a|s,\theta)}{\partial \theta} \Delta \theta + O(\Delta \theta^2)
$$
$$
= \pi(a|s,\theta) (1 + \alpha \psi(s,a)^T \tilde{w}) + O(\Delta \theta^2)
$$
$$
= \pi(a|s,\theta)(1+ \alpha f^\pi(s,a;w)) + O(\alpha^2)
$$
\hfill $\square$


\hfill \break

\end{document}